\documentclass[lettersize,journal]{IEEEtran}
\usepackage{amsmath,amsfonts}
\usepackage{bm}
\usepackage{algorithmic}
\usepackage{algorithm}
\usepackage{array}
\usepackage[caption=false,font=normalsize,labelfont=sf,textfont=sf]{subfig}
\usepackage{textcomp}
\usepackage{stfloats}
\usepackage{url}
\usepackage{booktabs}
\usepackage[table]{xcolor}
\usepackage{multirow}
\usepackage{xcolor}
\usepackage{graphicx}
\usepackage{hyperref}
\usepackage{cite}
\def\m{\mathfrak{m}}

\def\s{\boldsymbol{s}}

\def\a{\mathbf{a}}

\def\mA{\mathbf{A}}
\def\mW{\mathbf{W}}
\def\mC{\mathbf{C}}
\def\mS{\mathbf{S}}

\def\m{\mathfrak{m}}

\def\x{\bm{x}}

\def\y{\mathbf{y}}
\def\W{\mathbf{W}}
\def\m{\mathbf{m}}

\def\w{\textbf{w}}

\begin{document}

\title{Compress Any Segment Anything Model (SAM)}

\author{Juntong Fan, Zhiwei Hao, Jianqiang Shen, Shang-Ling Jui, Yi Zhang, Jing-Xiao Liao, Feng-Lei Fan$^*$

\IEEEcompsocitemizethanks{
\IEEEcompsocthanksitem Fenglei Fan (hitfanfenglei@gmail.com) is the corresponding author.
\IEEEcompsocthanksitem Juntong Fan and Feng-Lei Fan are with Frontier of Artificial Networks (FAN) Lab, Department of Data Science, City University of Hong Kong, Hong Kong, China SAR.
\IEEEcompsocthanksitem Zhiwei Hao and Jianqiang Shen are with the Data Storage Product Line, Huawei Technologies Co., Ltd. (haozhiwei@huawei.com; shenjianqiang@huawei.com)
\IEEEcompsocthanksitem Shang-Ling Jui is with Lagrange Mathematics and Computing Research Center (jui.shangling@huawei.com)
\IEEEcompsocthanksitem
Yi Zhang is with the School of Cyber Science and Engineering,
Sichuan University, Chengdu 610207, China, and also with the Key
Laboratory of Data Protection and Intelligent Management, Ministry
of Education, Sichuan University, Chengdu 610207, China (e-mail:
yzhang@scu.edu.cn).
\IEEEcompsocthanksitem
Jing-Xiao Liao is with Department of Industrial and Systems Engineering, The Hong Kong Polytechnic University, Hong Kong, SAR of China.

}
}

\markboth{Journal of \LaTeX\ Class Files,~Vol.~14, No.~8, August~2021}%
{Shell \MakeLowercase{\textit{et al.}}: A Sample Article Using IEEEtran.cls for IEEE Journals}


\maketitle

\begin{abstract}
Due to the excellent performance in yielding high-quality, zero-shot segmentation, Segment Anything Model (SAM) and its variants have been widely applied in diverse scenarios such as healthcare and intelligent manufacturing. Therefore, effectively compressing SAMs has become an increasingly pressing practical need. In this study, we propose \textit{Birkhoff}, a novel data‑free compression algorithm for SAM and its variants. Unlike quantization, pruning, distillation, and other compression methods, \textit{Birkhoff} embodies versatility across model types, agility in deployment, faithfulness to the original model, and compactness in model size. Specifically, \textit{Birkhoff} introduces a novel compression algorithm: Hyper-Compression, whose core principle is to find a dense trajectory to turn a high-dimensional parameter vector into a low-dimensional scalar. Furthermore, \textit{Birkhoff} designs a dedicated linear layer operator, HyperLinear, to fuse decompression and matrix multiplication to significantly accelerate inference of the compressed SAMs. Extensive experiments on 18 SAMs in the COCO, LVIS, and SA-1B datasets show that \textit{Birkhoff} performs consistently and competitively in compression time, compression ratio, post-compression performance, and inference speed. For example, \textit{Birkhoff} can achieve a compression ratio of 5.17$\times$ on SAM2-B, with less than 1\% performance drop without using any fine-tuning data. Moreover, the compression is finished within 60 seconds for all models. We have open-sourced our code in \url{https://github.com/Juntongkuki/Birkhoff-Model-Compression.git} for readers' free use.
\end{abstract}


\section{Introduction}
\IEEEPARstart{I}{N} computer vision, Segment Anything Model (SAM) \cite{kirillov2023segment} has emerged as a dominant solution for a wide range of segmentation tasks. Extensive independent evaluations confirm that SAM could achieve high-quality segmentation masks through natural interactions, \textit{i.e.}, clicks, bounding boxes, or text prompts. SAM's core objective is zero-shot segmentation, enabling segmentation of arbitrary objects without task-specific training. Building on SAM's significant success, as Table \ref{info-sams} shows, numerous variants like Med-SAM \cite{zhou2024medsam} and SAM-HQ \cite{ke2023segment} have been developed. They employ novel architectures and fine-tuning strategies to enhance efficacy in specific domains or boost segmentation performance further. SAM and its variants have been widely applied in diverse scenarios such as healthcare and intelligent manufacturing. However, SAM and many variants are notoriously large-scale, hindering efficient deployment for users with limited computational resources. For instance, MobileSAMv2 \cite{zhang2023mobilesamv2}, a compact SAM variant, still comprises 641 million parameters and requires 2.37 GB of storage. Deploying such large models presents significant challenges in resource-constrained settings, such as autonomous vehicles and smartphones. Consequently, effectively compressing SAM and its variants becomes an increasingly pressing practical need.

\begin{table}[ht]
    \begin{center}
    \caption{A summary of several mainstream SAM variants}
    \vspace{-0.1cm}
    \scalebox{0.75}{
    \setlength{\tabcolsep}{1mm}{
    \begin{tabular}{ccccc}
    \hline
    \textbf{Model} & \textbf{Pre-training Data} & \textbf{File Size} & \textbf{Parameters} & \textbf{linear params(\%)}\\
    \hline
    SAM-B\cite{kirillov2023segment} & SA-1B & 357MB & 94M & 94.84\% \\
    SAM-L\cite{kirillov2023segment} & SA-1B & 1.16GB & 312M & 97.96\% \\
    SAM-H\cite{kirillov2023segment} & SA-1B & 2.38GB & 641M & 98.76\%\\
    SAM-HQ-Tiny\cite{ke2023segment} & SA-1B + HQSeg-47K & 40.5MB & 11M & 86.12\%\\
    SAM-HQ-B\cite{ke2023segment} & SA-1B + HQSeg-47K & 361MB & 95M & 93.92\%\\
    SAM-HQ-L\cite{ke2023segment} & SA-1B + HQSeg-47K & 1.16GB & 314M & 97.58\%\\
    SAM-HQ-H\cite{ke2023segment} & SA-1B + HQSeg-47K & 2.39GB & 643M &  98.53\% \\
    SAM2-Tiny\cite{ravi2024sam} & SA-1B & 148MB & 39M & 97.58\% \\
    SAM2-Small\cite{ravi2024sam} & SA-1B & 175MB & 46M & 97.91\% \\
    SAM2-Base\cite{ravi2024sam} & SA-1B & 308MB & 81M & 98.64\% \\
    SAM2-Large\cite{ravi2024sam} & SA-1B & 856MB & 224M &  99.36\% \\
    MobileSAM\cite{zhang2023mobilesamv2} & 1\%SA-1B & 38.8MB & 10M &  88.56\% \\
    MobileSAMv2(ViT-H)\cite{zhang2023mobilesamv2} & 1\%SA-1B & 2.37GB & 641M & 98.76\% \\
    EdgeSAM\cite{zhou2023edgesam} & 1\%SA-1B & 37.0MB & 10M & 41.19\% \\
    EdgeSAM-RPN\cite{zhou2023edgesam} & 3\%SA-1B & 37.0MB & 10M & 41.19\% \\
    EfficientSAM-Ti\cite{xiong2024efficientsam} & SA-1B+IN & 39.0MB & 10M & 90.72\% \\
    EfficientSAM-S\cite{xiong2024efficientsam} & SA-1B+IN & 100MB & 26M & 95.41\% \\
    TinySAM\cite{shu2025tinysam} & SA-1B & 38.8MB & 10M & 88.56\%\\
    MedSAM\cite{ma2024segment} & FLARE & 358MB & 94M & 95.90\%\\
    \hline
    \end{tabular}}}
    \label{info-sams}
    \end{center}
    \vspace{-0.4cm}
\end{table}

We think that an ideal compression solution for SAMs should exhibit four key characteristics: \textbf{versatility} across model types, \textbf{agility} in model deployment, \textbf{faithfulness} to the original model, and \textbf{compactness} in model size. (i) The philosophy of SAMs lies in the universality, compression methods shall not be restricted to specific model architectures either. This universality enhances compression efficiency and facilitates the transfer of diverse SAM variants into massive users. (ii) The growing adoption of “model factory" paradigms in cloud computing necessitates parsimonious model handling. Compression must therefore be sufficiently rapid to enable efficient batch processing, which can significantly boost the overall compression throughput. (iii) Compressed models shall closely preserve the original SAM's capabilities. Given the substantial costs involved in training foundational SAMs, a large performance loss is unacceptable. Consequently, compression should not require model retraining. While model open-sourcing is common, accessing the data engine, \textit{e.g.}, as with Med-SAM, is often prohibited, and datasets prepared specifically for compression are typically limited. Retraining risks cumbersome procedures, catastrophic forgetting~\cite{catastrophicforgetting} (undermining SAM's core capabilities), and the potential introduction of unintended biases. (iv) Provided faithfulness is maintained, the compression solution shall achieve a high compression ratio to enable viable deployment in extremely resource-constrained environments.

Four principal model compression techniques are commonly employed: pruning, quantization, knowledge distillation, and low-rank decomposition. The compression ratios by pruning and quantization methods are hard to scale. Studies have shown that under the constraint of model fidelity, pruning can typically achieve only a compression ratio of 2 to 4 times \cite{zhu2024survey}. The theoretical lower bound for quantization is 1-bit representation, but subject to the severe perturbation to model parameters and the difficulty of fine-tuning. To the best of our knowledge, achieving a reliable and performant lower-than-INT4 quantization has been a challenging task
for industry and academia so far. Therefore, low-bit quantization remains impractical for models like SAM. Moreover, low-rank decomposition approximates weight matrices via low-rank representations but faces critical limitations. First, the method necessitates extensive tuning for convergence due to the non-convex optimization landscape. Second, low-rank decomposition struggles with standard convolution kernels which are small, limiting its applicability as a universal model compression technique \cite{li2023model}. As for knowledge distillation, it involves extensive training; therefore, we exclude it from our discussion. 

\begin{figure}[htb]
    \centering
    \includegraphics[width=0.8\linewidth]{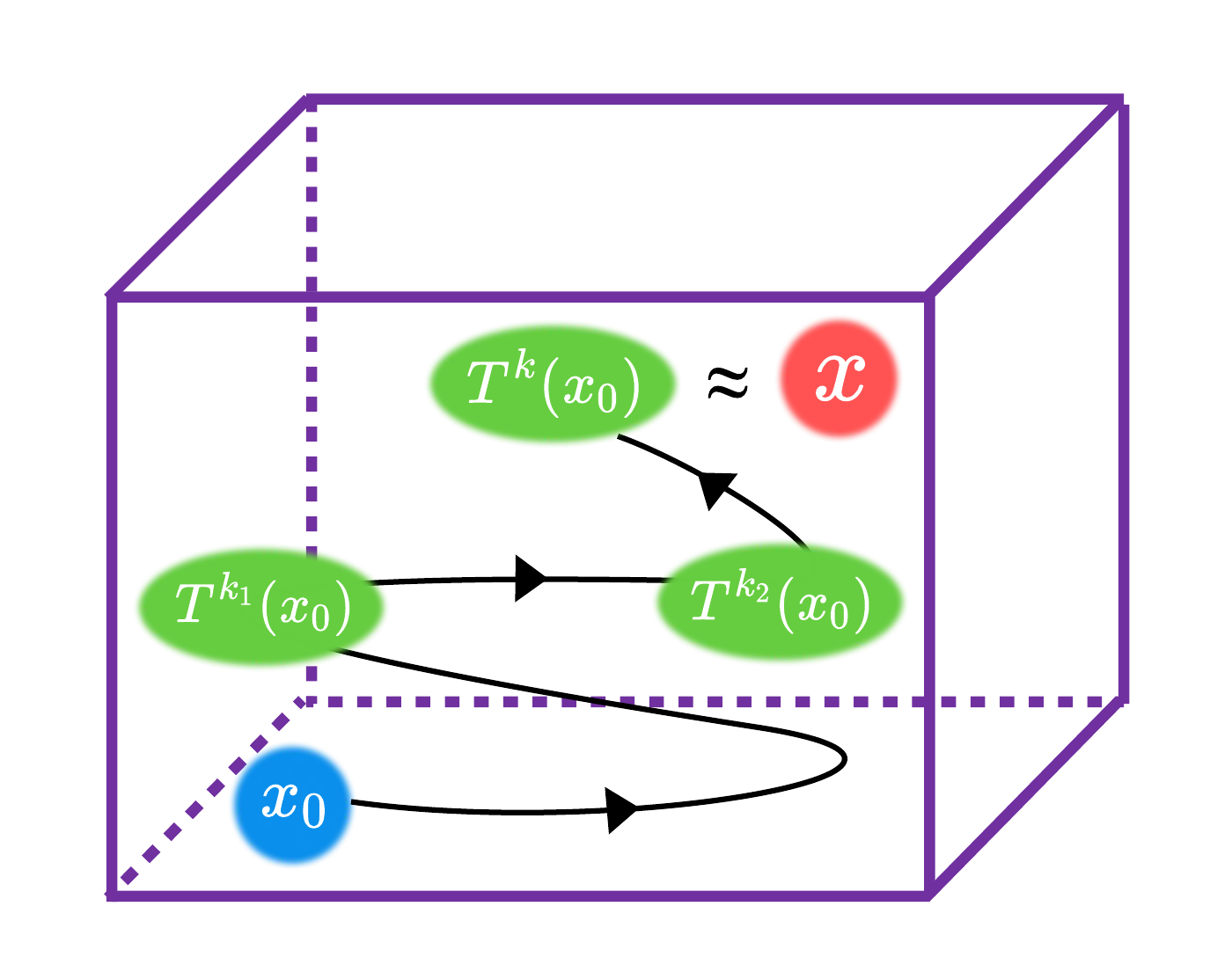}
    \caption{The basic principle of the Hyper-Compression, where $ \x \in \mathbb{R}^N $ denotes the target high-dimensional point. Some transformation $ T $ and the initial point $ \x_0 $ establish the existence of a scalar $ k \in \mathbb{N}^+$ for an arbitrary $\x$ such that $ \x $ can be approximately represented as $ T^k(\x_0) $. Thus, we can compress $\x$ into $k$. This figure is reproduced based on Figure 2 in \cite{fan2024hyper}.}.
    \label{fig:hyper}
\end{figure}

Earlier, our group proposed Hyper-Compression, which is a novel model compression technique that leverages trajectory density to transform the high-dimensional parameter compression into the representation of a low-dimensional dynamic system. As illustrated in Figure \ref{fig:hyper}, there exists a dynamic system such that the one-dimensional trajectory induced by some initial point can fill the high-dimensional space. Here, “filling the high-dimensional space” means that the trajectory can always arrive in the arbitrarily small neighborhood of any point. Thus, Hyper-Compression establishes a deterministic mapping between the one-dimensional trajectory parameter $ k $ (composition times or trajectory length) and the high-dimensional target vector $ \mathbf{x} \in \mathbb{R}^N $: $\mathbf{x}\approx T^k(\mathbf{x}_0)$. This potentially enables the compression of $ \mathbf{x} $ into the scalar $ k $: $\mathbf{x} \to k$, while decompressing $ \mathbf{x} $ from $ k $ follows the composition of transformation: $ T^k(\mathbf{x}_0) \approx \mathbf{x}$. As Table \ref{tab:principle} shows, Hyper-Compression enjoys a novel mechanism for model compression that is fundamentally different from quantization, pruning, and low-rank decomposition.

\begin{table}[htbp]
  \centering
  \caption{Hyper-compression searches a low-dimensional point for a high-dimensional parameter vector from the perspective of dense trajectory, which embraces an essentially different mechanism from the current three mainstream data-free model compression methods.}
  \scalebox{1.1}{
  \begin{tabular}{lc}
  \hline
      Method & Core Principle   \\
     \hline
  Quantization &  Low-precision \\
         Pruning &  Sparsity  \\
    Low-rank Decomposition  & Low-rank \\
    \textbf{Hyper-compression} & \textbf{Low-dimensionality} \\
    \hline
    \end{tabular}}%
    \label{tab:principle}
\end{table}

In this work, based on Hyper-Compression, we propose the \textit{Birkhoff} algorithm: a universal, data-free, fast, and high-accuracy-compression-ratio model compression algorithm. The algorithm is named after the great mathematician George David Birkhoff, who made pioneering contributions in dynamic systems. \textit{Birkhoff} introduces Hyper-Compression into the problem of SAM compression and makes major modifications to render Hyper-Compression infer much faster. Specifically, \textit{Birkhoff} designs HyperLinear, a dedicated linear layer operator, to fuse decompression and matrix multiplication, in support of GPU acceleration. As a result, a single access to parameters can retrieve a parameter vector simultaneously, thereby enhancing both the L2 cache hit rate and data transmission efficiency. Furthermore, we conduct localized rapid compression and agile deployment experiments using \textit{Birkhoff} across 18 SAM variants, with comprehensive evaluations spanning compression time, compression ratio, post-compression model performance, and inference speed. Experimental results on image segmentation tasks using three benchmark datasets: COCO\cite{lin2014microsoft}, LVIS\cite{gupta2019lvis}, and SA-1B\cite{kirillov2023segment} demonstrate that, in most cases, our method achieves a compression ratio greater than 4$\times$, which is the theoretical maximum attainable through INT8 quantization, while keeping performance degradation within 1\%. Moreover, the compression process is highly efficient, taking no more than 60s in all cases. In summary, our main contributions are threefold:

\begin{itemize}

   \item Hyper-compression is a novel mechanism for model compression, which holds the great potential in addressing the gap between computation and hardware resource. To the best of our knowledge, our work is the first to use it in compressing SAMs, fully verifying its broad applicability from language models to vision models. 

    \item We propose \textit{Birkhoff}, an innovative efficient model compression framework, along with a custom-designed CUDA operator, HyperLinear, fusing the decompression and matrix multiplication. With HyperLinear, \textit{Birkhoff} can significantly improve the inference speed of the compressed model to the level of the original model.

    \item We comprehensively evaluate \textit{Birkhoff} from the compression ratio, compression time, post-compression inference speed, and accuracy degradation on 18 SAM variants. Furthermore, we also compare the post-compression performance of our method with 8 state-of-the-art model compression approaches. Results show that our method consistently delivers competitive compression performance.
\end{itemize}

\section{Related Work}

\subsection{Data-Free Model Compression}
Most model compression methods require fine-tuning. However, data-free model compression approaches have increasingly attracted attention from the research community due to their better user-friendliness.

\subsubsection{Post-training pruning}
Pruning is a method to achieve sparsity in neural networks by removing redundant or unimportant synapses, neurons, layers, or even entire blocks. SparseGPT\cite{frantar2023sparsegpt} formulates the pruning problem as a series of extremely large-scale sparse regression tasks, which are then efficiently solved using a novel approximate sparse regressor. SynFlow\cite{tanaka2020pruning} addresses the issue of layer collapse in gradient-based pruning methods at initialization. Inspired by the lottery ticket hypothesis \cite{frankle2018lottery}, SynFlow designs the algorithm to find the effective sub-network that can preserve the total flow of synaptic strengths. Through dynamic score adjustments, SynFlow ensures a balanced parameter distribution across layers, thereby avoiding over-pruning in large layers. AutoDFP\cite{li2024autodfp} establishes automatic data-free pruning and reconstruction based on the assumption that the loss of information can be partly compensated by retaining essential information from similar channels. Specifically, the similarity information of channels for each layer can guide a reinforcement learning agent in the pruning and reconstruction of the network.

\subsubsection{Post-training quantization (PTQ)}PTQ\cite{huang2024billm} applies quantization after the training phase without further updating its parameters. MinMax\cite{jacob2018quantization} proposes a linear quantization scheme that enables efficient computation by performing arithmetic operations solely on integer values. 
This method models the mapping from a real-valued weight or an activation $r$ to a quantized integer $q$ as an affine transformation, parameterized by a scale factor $S$ and a zero-point $Z$. 
Denoting $[\beta, \alpha]$ the range of representable real values chosen for quantization and $b$ the bit-width of the signed integer representation, this transformation is defined as follows:
\begin{equation}
\left\{
\begin{aligned}
    q &= \texttt{clip}(\texttt{round}(S \times r + Z), -2^{b-1}, 2^{b-1}-1) \\
    S &= \frac{2^b-1}{\alpha - \beta} \\
    Z &= -\texttt{round}(\beta \times S)-2^{b-1}
\end{aligned}
\right.,
\end{equation}
where $\texttt{round}()$ means the nearest integer, and $\texttt{clip}()$ is a truncation function. Percentile~\cite{wu2020integer} sets the quantization range $\alpha$ and $\beta$ to a specified percentile of the absolute value distribution observed in model weights or activations. For instance, the $99\%$ percentile corresponds to clipping 1\% of the largest magnitude values. OMSE~\cite{choukroun2019low} employs multiple low-precision quantized tensors to replace mixed-precision quantization for better approximation, by formulating the quantization problem as a constrained mean squared error (MSE) minimization problem. Let $n$ integers $\{p_i\}^n_{i=1}$ denote the desired precision of the quantized tensors and a given tensor $T$, the optimization target can be defined as follows:
\begin{equation}
\begin{aligned}
\min_{\alpha_i, \tilde{T}^i} &\quad \left\| T - \sum_{i=1}^{n} \alpha_i \tilde{T}^i \right\|_F^2, 
~~\text{s.t.}~  \alpha_i \in \mathbb{R},\ \tilde{T}^i \in \mathbb{Z}_{p_i}^{c \times w \times h},
\end{aligned}
\end{equation}
where $\tilde{T}^i$ is the $i$-th quantized tensor, $\alpha_i$ is the $i$-th scaling factor, and $c,w,h$ are the sizes of the tensor. ZeroQ~\cite{cai2020zeroq} leverages a novel sensitivity analysis method based on the KL divergence to measure each layer's tolerance capability to the parameter loss. The tolerance measurement uses the synthetic distilled data to match the statistics of the batch normalization layers. Therefore, it can automatically select a suitable quantization configuration for mixed-precision according to a pareto frontier optimization.

\subsubsection{Post-training low-rank decomposition}
Low-rank decomposition \cite{oseledets2011tensor, zhao2016tensor} involves approximating model parameter matrices or tensors by decomposing them into low-rank ones: 
\begin{equation}
A(i_1, i_2, \ldots, i_d) = \sum_{\alpha = 1}^{r} U_1(i_1, \alpha) U_2(i_2, \alpha) \cdots U_d(i_d, \alpha),
\end{equation}
which can lead to a more compact representation. The research about data-free low-rank decomposition used for model compression is scarce. \cite{peng2024data} applied Rank-k approximation to compress large language models. Specifically, it proposes the so-called joint Rank-k approximation~\cite{peng2024data} by drawing the insight that performing Rank-k approximation solely on individual layers overlooks the interaction between related modules. Therefore, the joint Rank-k approximation decomposes the merged linear layers/attention modules to retain more useful information as well as compress more.

\subsection{Compression of SAM}
Knowledge Distillation (KD)\cite{hinton2015distilling} is a widely adopted method for compressing SAMs, which attempts to employ the knowledge from a cumbersome model to train a small yet powerful model. MobileSAM\cite{zhang2023faster} proposes a decoupled distillation method that enables knowledge distillation from a heavy image encoder to a lightweight one. Specifically, MobileSAM divides the knowledge distillation task into two subtasks: image encoder distillation and mask decoder fine-tuning. It directly distills a compact image encoder from the original SAM without relying on the original mask decoder, and subsequently fine-tunes the mask decoder using the frozen lightweight image encoder to further enhance performance. To overcome the bottleneck that task-agnostic encoder distillation fails to capture the full knowledge embodied in the original SAM, EdgeSAM\cite{zhou2023edgesam} proposes a dynamic prompt-in-the-loop strategy to include both the prompt encoder and mask decoder in the distillation process, instead of encoder-only KD. EdgeSAM can accurately capture the intricate dynamics between user input and mask generation. For maintaining the strong zero-shot performance, TinySAM\cite{shu2025tinysam} proposes a full-stage KD method with hard prompt sampling and hard mask weighting strategy to distill a lightweight student model. Based on an observation that masks of SA1B dataset could be extremely diverse and fine-grained in a single image, the hard mask weighting is proposed to integrate an additional value called mask hardness ($H_i$) into the distillation loss. Furthermore, the hard prompt sampling is also introduced to further ease the learning process, thereby concentrating the training samples in the difficult area for prediction. 

Quantization is another common technique for compressing SAMs. For example, PTQ is performed on the above-mentioned TinySAM \cite{shu2025tinysam} to further reduce the memory demand. PTQ4SAM\cite{lv2024ptq4sam} addresses the unique bimodal distribution of parameters, the phenomenon observed in SAM. By simultaneously incorporating the sign factor into both the query and key linear layers, the method effectively transforms the bimodal distribution into a normal distribution. This transformation results in a distribution range that exhibits significantly improved quantization performance. 

We argue that quantization followed by fine-tuning and knowledge distillation are more prone to introduce the catastrophic forgetting in SAMs because involving training inherently risks overwriting or distorting the original learned representations. Especially when the fine-tuning data is limited or unrepresentative, training causes the model to prioritize quantization robustness over previously acquired knowledge. Due to the strong data engine used in training SAM, it is hard to guarantee that the self-collected dataset is representative. In contrast, direct compression without fine-tuning applies a static transformation without updating weights, thereby preserving the model's original knowledge structure well.

\subsection{Introduction of Hyper-Compression} \label{hyper_theory}
Based on ergodic theory, Hyper-Compression can both lower the parametric precision and cut down the number of parameters at the same time, resulting an effective compression. Mathematically, a hyperfunction $h(\theta;n)$ is used to encode and decode the relationship between the locations and weights of a given set $\{w_n\}_{n=1}^N$ split from the target network. How to select a proper hyperfunction remains an open problem. Hyper-Compression draws on knowledge from ergodic theory, which characterize the conditions under which the trajectory of a low-dimensional dynamical system fills up a higher-dimensional space \cite{cornfeld2012ergodic}. When such a property holds, one can, from an engineering standpoint, exploit low-dimensional curves to approximate arbitrary points in the high-dimensional space. Consequently, a larger number of model parameters $\{w_n\}_{n=1}^N$ can be represented through a smaller number of parameters $\theta$ by explicitly compressing the target points via the trajectory length of the chosen dynamical system. In other words, for each $w_n$, $\forall \epsilon > 0$, $\exists~ \theta^* > 0$ such that 
\begin{equation}
    |w_n-h(\theta^*; n)| < \epsilon,
\label{related_hyper}
\end{equation} where the storage memory usage for $\theta^*$ is much less than that for $\{w_n\}_{n=1}^N$. Table \ref{tab:principle} summarizes that Hyper-Compression enjoys an essentially different mechanism from the current three mainstream data-free model compression methods.

Hyper-Compression shares the same spirit with the vector quantization (VQ) \cite{gray1984vector}. This technique also can establish a mapping from a $ k $-dimensional vector in the space $ \mathbb{R}^k $ to a common vector $\mathbf{y}_i$ through clustering. Each vector $ \mathbf{y}_i $ is referred to as a codeword, and the set of all the codewords is called a codebook. For each codeword $ \mathbf{y}_i $, there exists a corresponding nearest-neighbor region known as a Voronoi region, which is formally defined as
\begin{equation}
    V_i = \{ \x \in \mathbb{R}^k : ||\x - \y_i || \leq || \x - \y_j ||, \forall j \neq i \}.
\end{equation}
Moreover, the Voronoi regions constitute the entire space $\mathbb{R}^k$ and do not overlap with each other: 

\begin{equation}
\left\{
\begin{aligned}
    \quad \bigcup^N_{i=1} V_i &= \mathbb{R}^k \\
    \bigcap^N_{i=1} V_i &= \emptyset, \quad i \neq j. \\
\end{aligned}
\right.
\end{equation}
The Euclidean distances between each input vector and all codewords in the codebook need to be computed. The index of the codeword that has the shortest distance from the given input vector is determined. Given $m$ $D$-dimensional vectors, if we want to cluster them into $K$ classes, the computational complexity is $\mathcal{O}(mDK)$. Therefore, constructing a suitable codebook is also of $\mathcal{O}(mDK)$ complexity, which is overly high for large models with a great amount of parameters. Thus, this method cannot be easily used into LLM compression problem. This is because large models have a tremendous amount of parameters. In contrast, Hyper-Compression selects a codebook along a $1$-dimensional trajectory, which can avoid the clustering. 



\begin{figure*}[ht]
    \centering
    \includegraphics[width=1\linewidth]{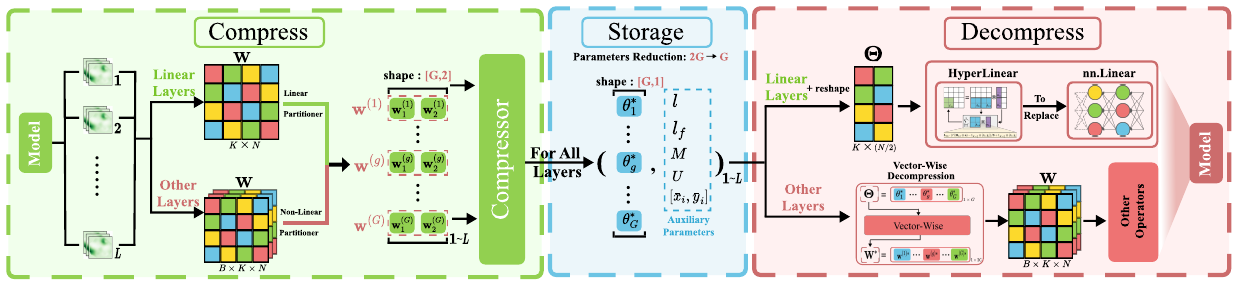}
    \caption{An overview of our \textit{Birkhoff} compression framework. First, model parameters are subjected to per-tensor compression and then stored with auxiliary global parameters. Then, the compressed result of each tensor is packed and stored with its corresponding auxiliary global parameters. Lastly, parameters from a model’s linear layers are loaded into the HyperLinear operator layer by layer for inference.}
    \label{overview}
\end{figure*}

\section{Methodology}

The framework of \textit{Birkhoff} is illustrated in Figure~\ref{overview}. During the compression stage, \textit{Birkhoff} employs Hyper-Compression to perform layer-wise compression on SAM parameters $\w$, achieving at least a 50\% reduction in parameters $\boldsymbol{\theta^*} < \w$, while maintaining acceptable error levels. In the decompression stage, we introduce HyperLinear, which combines parameter decompression with matrix multiplication to replace traditional linear layers in neural networks. With the acceleration of CUDA, this approach significantly achieves much faster model inference.

\textit{Birkhoff} satisfactorily addresses the major weakness of Hyper-Compression in inference time. Unlike traditional Hyper-Compression, which follows a sequential ``decompress-then-compute'' approach, \textit{Birkhoff} accelerates decompression through tensor-parallel computation. This innovation dramatically speeds up inference, particularly designed for linear layers that contain the majority of parameters in SAMs.

In this section, we first describe Hyper-Compression algorithm in detail, as the necessary preliminary information. Subsequently, we explain how to integrate decompression with a matrix product, which constitutes the mathematical foundation of HyperLinear.


\subsection{Algorithmic Details of Hyper-Compression} \label{hc_details}

Without loss of generality, Hyper-Compression assumes $w_n \in [0,1], n=1,\cdots,N$. For computational simplicity, the hyperfunction $h(\theta;n)$ is cast as $\tau(\theta\cdot a_n)$, where $\tau(z) = z - \lfloor z \rfloor$, $a_n = \frac{1}{\pi+n}$, and {$\theta > 0$}. When $\theta$ goes from 0 to $\infty$, $[\tau(\theta\cdot a_1),\tau(\theta\cdot a_2),\cdots,\tau(\theta\cdot a_N)]$ forms a trajectory in $[0,1]^N$. Since $a_1, \cdots, a_N$ are irrationally independent, the trajectory never overlaps with itself, and therefore eventually fills the entire space $[0,1]^N$. Furthermore, due to the density of trajectory in $[0,1]^N$, given any $\{w_n\}_{n=1}^N$, there exists $\theta^*$ such that
\begin{equation}
    w_n \approx \tau(\theta\cdot a_n), ~~n=1,\cdots,N.
    \label{eqn:hc}
\end{equation}
Hyper-Compression uses Eq. \eqref{eqn:hc} to do model compression and focuses on $N=2$. 

Specifically, given a parameter matrix, Hyper-Compression first flattens the matrix into a one-dimensional vector $\w = [w_1, w_2, \cdots, w_L]$, where $L$ is the total number of elements in $\w$, and then splits $\w$ into groups $[\w^{(1)}, \cdots, \w^{(g)}, \dots, \w^{(G)}]$, where $\w^{(g)}=[w_{2(g-1)+1}, w_{2(g-1)+2}]$ is of $\mathbb{R}^{1\times 2}$. In practice, if $N$ is not divisible by $2$, we calculate the average of $[w_1, w_2, \cdots, w_L]$ as $w_{L+1}$, and add it to the end of $[w_1, w_2, \cdots, w_L]$. As a result, for $w^{(g)}_i, i=1,2$, $\forall \epsilon > 0$, there exists a value $\theta^*_g > 0$, so that 
\begin{equation}
    |w^{(g)}_i-\tau(\theta^*_g \cdot a_i )| < \epsilon.
    \label{0_1_leq}
\end{equation} 
Thus, Hyper-Compression compresses $\w^{(g)}$ into $\theta^*_g$. Accordingly,  $\w=[\w^{(1)}, \cdots, \w^{(g)}, \cdots, \w^{(G)}]$ can be compressed into $\boldsymbol{\theta^*} =[\theta^*_1, \cdots, \theta^*_g, \cdots, \theta^*_G]$. Let $w^{(g)*}_i=\tau(\theta^*_g \cdot a_i )$, we have 
\begin{equation}
    w^{(g)}_i \approx w^{(g)*}_i.
    \label{compress_8}
\end{equation}
For convenience, decompressing the whole vector $[\theta^*_1, \cdots, \theta^*_g, \cdots, \theta^*_G]$ can be done in the vector-wise manner:
\begin{equation}
\begin{aligned}
    \w^* &= [\w^{(1)*}, \cdots, \w^{(g)*}, \cdots, \w^{(G)*}] \\
    &= \tau(\boldsymbol{\theta^*} \otimes \boldsymbol{a}),
\end{aligned}
\label{0_1_eq}
\end{equation}
where $\w^{(g)*}=[w^{(g)*}_1, w^{(g)*}_2]$, $\boldsymbol{a}=[a_1,a_2]$, and $\otimes$ means the Kronecker product.

\begin{figure}[htb]
    \centering
    \includegraphics[width=1\linewidth]{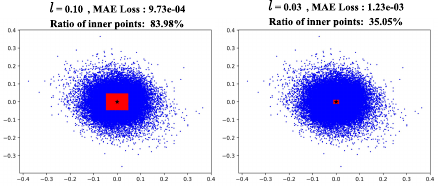}
    \caption{The relationship between the parameter $l$ and the proportion of inner points.}
    \label{l_exams}
\end{figure}

Furthermore, to adapt the theory into practical model compression scenarios, Hyper-Compression systematically designs four engineering twists:

 \textit{1)} The parameters of a given model may not always fall within the range $[0,1]^2$. Therefore, the unit square $[0,1]^2$ is replaced with a more flexible box $[\bar{x}_i-\frac{l}{2},\bar{x}_i+\frac{l}{2}]\times[\bar{y}_i-\frac{l}{2},\bar{y}_i+\frac{l}{2}]$, a square with the length $l$ and centered at $[\bar{x}_i, \bar{y}_i]$, where $[\bar{x}_i, \bar{y}_i]$ is the centroid of two-dimensional points $\w^{(1)}, \w^{(2)}, \dots, \w^{(G)}$. Using $[\bar{x}_i-\frac{l}{2},\bar{x}_i+\frac{l}{2}]\times[\bar{y}_i-\frac{l}{2},\bar{y}_i+\frac{l}{2}]$ instead of $[0,1]^2$ actually means that two-dimensional points are translated not into the latter but into the former. $l$ is the size of the box determined based on the parameter distribution. The selection of $l$ should ensure that most points in $[\w^{(1)},...,\w^{(G)}]$ are placed in the box $[\bar{x}_i-\frac{l}{2},\bar{x}_i+\frac{l}{2}]\times[\bar{y}_i-\frac{l}{2},\bar{y}_i+\frac{l}{2}]$, and the other points outside the box will be scaled into the box before compression. As shown in Figure \ref{l_exams}, two figures illustrate the relationship between the parameter $l$ and the proportion of inner points, where the points inside the box are colored by red while the points outside the box are colored by blue, which contributes to the MAE loss of a layer parameter. When $l=0.1$, approximately $83.98\%$ of the points of $\{ \w^{(0)},...,\w^{(G)} \}$ fall within the inner square region, resulting in an MAE loss of $9.73\times10^{-4}$. In contrast, when $l=0.03$, only $35.05\%$ of the points are located within the box, and the MAE loss increases significantly to $1.23\times10^{-3}$.

 \textit{2)} Hyper-Compression needs to scale points into the box $[\bar{x}_i-\frac{l}{2},\bar{x}_i+\frac{l}{2}]\times[\bar{y}_i-\frac{l}{2},\bar{y}_i+\frac{l}{2}]$. To deal with outliers, it adopts different scaling factors for different points. Specifically,  all the points $\w^{(1)}, \w^{(2)}, \dots, \w^{(G)}$ are divided into $M$ cetegories based on their distances from the centroid $[\bar{x}_i, \bar{y}_i]$. For example, if the distance between the centroid and $\w^{(g)}$ is $l_g$, its category $m^{(g)}$ is computed as follows:
    \begin{equation}
    m^{(g)} = 
    \left\{
    \begin{array}{ll}
    0 & \text{if } l_g \leq l/2  \\
    \lceil \frac{M \times (2 \times l_g-l)}{2 \times l_f-l} \rceil & \text{if } l_g > l/2 ,
    \end{array}
    \right.
    \end{equation}
    where $l_f$ is the global farthest distance from $[\bar{x}_i, \bar{y}_i]$, and $M$ is a hyperparameter. Then, for the points under the $m$-th category, if $m$ is large, the corresponding scaling factor is also large accordingly. Mathematically, the scaling factor $s_m$ is defined as 
    \begin{equation}
        s= \frac{l}{l + \frac{m}{M} \cdot (2l_f - l)}.
        \label{scale_factor}
    \end{equation}
    Vector-wise, $[\w^{(1)}, \w^{(2)}, \dots, \w^{(G)}]$ uses the scaling factor vector $\boldsymbol{s}=[s^{(1)}, \cdots, s^{(g)}, \cdots, s^{(G)}]$.

    \textit{3)} In algorithmic implementation, to save the storage space, Hyper-Compression discretizes $\theta$ in Eq. \eqref{eqn:hc} as $\Delta\cdot\theta$, where $\Delta$ is a common interval, and $\theta$ is an integer. Here, we slightly abuse the notation $\theta$ for succinct expression. To further reduce the storage, we don't want $\theta$ to be huge. Therefore, we set an integer $U$ as the upper bound of $\theta$. Then, since $\theta$ is only cast from ${0,1,\cdots,U}$, searching $\theta$ turns into judging which integer from $\{0,1,\cdots,U\}$ can lead to the smallest error based on Eq. \eqref{eqn:hc}. Now, let us illustrate the roles of different hyper-parameters. A smaller $U$ leads to a higher compression ratio. $U$ controls how many points we sample from the one-dimensional trajectory. A smaller $U$ may cause a higher approximation error.
    
    \textit{4)} $\a=[a_1,a_2]=[\frac{1}{\pi+1}, \frac{1}{\pi + 2}]$ can be adjusted according to the distribution of $[\w^{(1)}, \w^{(2)}, \dots, \w^{(G)}]$ to further minimize the approximation error. If the distribution is uniform, we can set $\a$ as
    \begin{equation}
    \begin{aligned}
        \a & = \frac{[\frac{l}{U},l]}{\sqrt{(\frac{l}{U})^2+l^2}}
    \label{vector_m1}
    \end{aligned}
    \end{equation}
    to also make $\tau'(\theta\cdot \a), \theta=0,1,\cdots,U$ distribute uniformly.

According to the above twists, an outlier point $\w$ under $m$-th category, the final compression of $\w$ is given by the following formula:
\begin{equation}
    \w \approx \frac{\tau'(\theta \cdot \a)-[\bar{x}_i, \bar{y}_i]}{\s}+[\bar{x}_i, \bar{y}_i], ~~n=1,\cdots,N,
    \label{eq_opt_compress}
\end{equation}
where $\tau'([x,y]): \mathbb{R}^2 \to \mathbb{R}^2: = f([x,y]) + [\bar x_i, \bar y_i] - [\frac{l}{2}, \frac{l}{2}]$, and $f(x) = x \bmod U$.

As a result, the current formulation is different from that in Eq. \eqref{0_1_eq} as follows:
\begin{equation}
\begin{aligned}
    &\w^*= [\w^{(1)*}, \cdots, \w^{(g)*}, \cdots, \w^{(G)*}]= \\
    & (\tau'(\boldsymbol{\theta}^*\otimes \boldsymbol{a})-[1,\cdots,1] \otimes [\bar{x}_i, \bar{y}_i]) / \boldsymbol{s} + [1,\cdots,1] \otimes [\bar{x}_i, \bar{y}_i].
\end{aligned}
\label{scale_eq}
\end{equation}

From Eq.~\eqref{scale_eq}, the decompression step calculates all scaling factors $\mathbf{s}$ by Eq.~\eqref{scale_factor}. To minimize storage overhead, we can avoid storing $\mathbf{s}$ by just saving $\boldsymbol{\theta^*}+ \m\cdot U$, where $\m=[m_1, \cdots, m_g, \cdots, m_G]$ are the categories associated with $\mathbf{s}$. Thus, in decompression, $\m$ can be obtained as
\begin{equation}
    m_g = \lfloor (\theta^*_g+ m_g\cdot U) / U \rfloor.
\end{equation}
This equation holds true because $\theta_g^*$ is an integer no more than $U$. For enhancing understanding of readers, in Table \ref{tab:hphc}, we summarize the hyper-parameters and their implications in Hyper-Compression.

\begin{table}[htbp]
\centering
\caption{A summary of hyper-parameters in Hyper-Compression.}
\begin{tabular}{l|l}
\hline
Hyper-parameters           & 
Intepretation          \\ \hline
$l$ & the length of the square box \\ \hline
$M$ & the total number of categories   \\ \hline
$U$ & an integer upper bound of $\theta$ \\ \hline
$\a$ & the direction of trajectory \\ \hline
$\Delta$ & the step size of trajectory \\ \hline
\end{tabular}
\label{tab:hphc}
\vspace{-0.2cm}
\end{table}

\begin{figure*}[ht]
    \centering
    \vspace{-0.5cm}
    \includegraphics[width=0.8\linewidth]{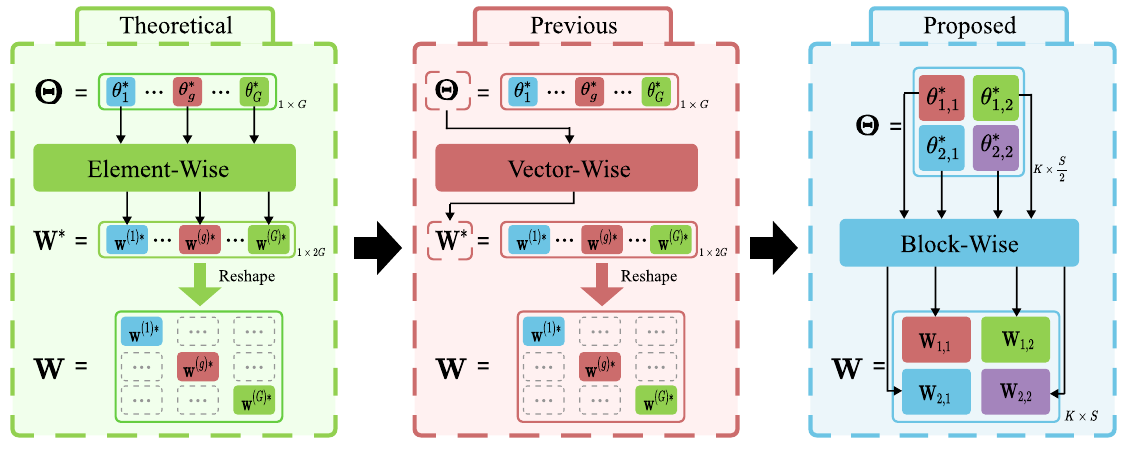}
    \caption{Technical evolution of decompression: from theoretical element-wise operation to vector-wise operation, and finally to our proposed block-wise one to facilitate operator fusion.}
    \label{wise_fig}
    \vspace{-0.4cm}
\end{figure*}

\subsection{Matrix Product in GPU Acceleration}

The core of neural network computation is the matrix product, whose speed affects the network inference time a lot. To accelerate matrix product, neural networks have adopted a GPU-parallelizable strategy in their linear layers. This strategy is "block-wise", which partitions a matrix into blocks for efficient parallel processing. For example, given three matrices that $\mA \in \mathbb{R}^{M \times K}$, $\mW \in \mathbb{R}^{K \times N}$, and $\mC=\mA \mW \in \mathbb{R}^{M \times N}$, The matrices $\mA$, $\mW$, and $\mC$ are partitioned into blocks ($\mA_{i,k}$, $\mW_{k,j}$, and $\mC_{i,j}$) according to the following formulas:
\begin{equation}
\left\{
\begin{aligned}
\mA_{i,k} &= \mA[i \cdot R - (R-1) : i \cdot R, \, k \cdot T-(T-1) : k \cdot T] \\
\mW_{k,j} &= \mW[k \cdot T - (T-1) : k \cdot T, \, j \cdot S-(S-1) : j \cdot S] \\
\mC_{i,j} &= \mC[i \cdot R - (R-1) : i \cdot R, \, j \cdot S-(S-1) : j \cdot S],
\end{aligned}
\right.
\end{equation}
where $\mA_{i,k} \in \mathbb{R}^{R \times T}$, $\mW_{k,j}\in \mathbb{R}^{T \times S}$, $\mC_{i,j} \in \mathbb{R}^{R \times S}$. $i \in \{1,\cdots,M/R\}$, $j \in \{1,\cdots,N/S\}$, and $k \in \{1,\cdots,K/T\}$, which means that the matrix $\mC$ can be computed through a parallel process of $R \times S$ blocks. Therefore, each block in $\mC$ can be calculated as follows:

\begin{equation}
\mC_{i,j} = \sum_{k=1}^{K/T} \mA_{i,k} \cdot \mW_{k,j}.
\end{equation}

\subsection{Hyperlinear:  Fuse decompression and matrix product}
\label{opt_alg}

The previous Hyper-Compression is vector-wise, which is incompatible with the matrix product in GPU acceleration that is “block-wise" operated. Therefore, Hyper-Compression needs to decompress the entire matrix first before doing the matrix product, which has large room for improvement. 

In the \textit{Birkhoff} algorithm, we design a “block-wise" compression and decompression, respectively, as shown in Figure \ref{wise_fig}.  Then, we propose Hyperlinear to fuse decompression and matrix product, which can greatly expedite the inference because linear layers including $Q, K, V$ attention modules usually dominate SAM. As shown in Table \ref{info-sams}, linear layers constitute $>95\%$ of parameters in SAM and variants like SAM-HQ, SAM-2, and MedSAM. The ratio of linear layers is relatively small in MobileSAM, EdgeSAM, TinySAM because they reduce the image encoder's parameters by knowledge distillation.

\subsubsection{Compression} Given a parameter matrix $\W \in \mathbb{R}^{K \times N}$, instead of flattening it, we split $\W$ into blocks of $1\times 2$ in each row, and the block in the $k$-th row is denoted as $\w$:
\begin{equation}
    \w^{(k,j)}=\W[k,2j-1:2j],
    \label{block_wise_eq}
\end{equation}
where $k \in \{ 1,\cdots,K \}$, and $j \in \{ 1,\cdots, N/2 \}$. Here, we assume that $N$ can be divided by $2$. Therefore, the compression of each $\w^{(k,j)}$ can be stored as one element $\mathbf{\Theta}[k,j]$ of a matrix $\mathbf{\Theta} \in \mathbb{Z}^{K \times \frac{N}{2}}$. If not, we need to augment the shape of $\W$ to $K \times (N+1)$ by padding a column that is donated as follows:
\begin{equation}
    \W[k,N+1] = \frac{2 \cdot\sum_{i=1}^{\frac{N-1}{2}}\W[k,2i]}{N-1},k=1,\cdots,K.
    \label{block_padding}
\end{equation}
This padding value is the mean of the second dimension across all points in the $k$-th row, such that this value does not affect the distribution of points significantly. There are three hyperparameters for the compression: $M$, $U$, and $l$. We consider using the mean absolute error (MAE) between the original tensor and decompressed lossy tensor, and select the hyperparameter configuration that can minimize the MAE loss. We design a compressor, which can automatically determine the best configuration for a given tensor, and then compress it, as shown in Algorithm \ref{pseudocode}. The compressor takes as input the parameter matrix $\W$ and three sets of alternative hyperparameters: $M$, $U$, and $l$.

\subsubsection{Decompression} Given $\mathbf{\Theta} \in \mathbb{Z}^{K \times (N/2)}$, as shown in Figure \ref{wise_fig}, we can recover $\W$ from $\mathbf{\Theta}$ as follows:
\begin{equation}
\left\{
\begin{aligned}
    s &= \frac{l}{l + \frac{\lfloor \mathbf{\Theta}[k,j] / U \rfloor}{M} \cdot (2l_f - l)} \\
    \a & = \frac{[\frac{l}{U},l]}{\sqrt{(\frac{l}{U})^2+l^2}} \\
    \W[k,2j-1:2j] &= (\tau'(\mathbf{\Theta}[k,j] \times \a)-[\bar{x}_i, \bar{y}_i]) / s + [\bar{x}_i, \bar{y}_i] \\
\end{aligned}.
\right.
\label{block_decom}
\end{equation}
As shown in Eq.~\eqref{block_decom}, decompressing $\mathbf{W}$ only requires a few auxiliary global parameters: $l$, $[\bar{x}_i, \bar{y}_i]$, $U$, $l_f$, and $M$, which are negligible compared to $\boldsymbol{\Theta}$. Therefore, they have no impact on the compression efficiency.

\begin{algorithm}[ht]
  \caption{Compress a parameter matrix.}
  \label{pseudocode}
  \begin{algorithmic}[1]
    \REQUIRE Parameter matrix $\mathbf{W}$; alternatives $M=\{M_i\}_{i=1}^{m}$; alternatives $U=\{U_j\}_{j=1}^{n}$; alternatives $l=\{l_k\}_{k=1}^{p}$
    \ENSURE Compression result $\boldsymbol{\Theta}$; A list of Auxiliary Parameters \textbf{Aux}

    \STATE If padding is needed, pad a column for $\mathbf{W}$ via Eq.~\eqref{block_padding}
    \STATE Split $\mathbf{W}$ and flatten into points $\{w^{(i)}\}_{i=1}^G$ via Eq.~\eqref{block_wise_eq}
    \STATE Compute center $[\bar x, \bar y]$ and global farthest distance $l_f$ of $\{w^{(i)}\}_{i=1}^G$
    \STATE least\_mae $\leftarrow \infty$;\quad $\boldsymbol{\Theta}\leftarrow \text{None}$;\quad \textbf{Aux}$\leftarrow \text{None}$

    \FOR{$i=1$ to $m$}
      \FOR{$j=1$ to $n$}
        \FOR{$k=1$ to $p$}
          \STATE Generate codebook CB
          \STATE mae\_loss $\leftarrow 0$
          \STATE Compress $\{w^{(i)}\}^G_{i=1}$ to $\boldsymbol{\theta}^*$ via Eq.~\eqref{eq_opt_compress}
          \STATE Decompress $\boldsymbol{\theta}^*$ to $\{w^{(i)*}\}^G_{i=1}$ via Eq.~\eqref{scale_eq}
          \STATE Compute mae\_loss between $\{w^{(i)*}\}^G_{i=1}$ and $\{w^{(i)}\}^G_{i=1}$
          \IF{mae\_loss $<$ least\_mae}
            \STATE least\_mae $\leftarrow$ mae\_loss
            \STATE $\boldsymbol{\Theta}\leftarrow \text{reshape}(\boldsymbol{\theta}^*)$
            \STATE \textbf{Aux}$\leftarrow [\,l_k,\,U_j,\,M_i,\,[\bar x,\bar y],\,l_f\,]$
          \ENDIF
        \ENDFOR
      \ENDFOR
    \ENDFOR

    \RETURN $\boldsymbol{\Theta}$, \textbf{Aux}
  \end{algorithmic}
\end{algorithm}

\subsubsection{Fusion} 
\begin{figure}[htb]
    \centering
    \includegraphics[width=0.9\linewidth]{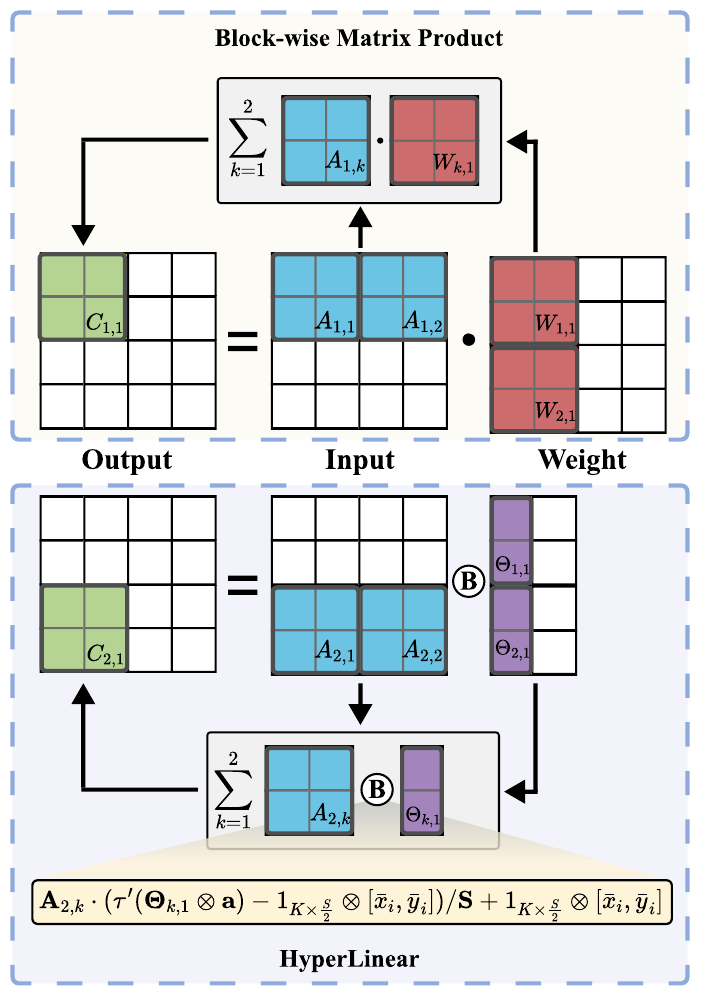}
    \caption{The comparison between the block-wise matrix multiplication and HyperLinear operation as Eq. \eqref{eq_birkhoff} .}
    \label{HyperLinear_fig}
\end{figure}

We can fuse the matrix product with the block-wise decompression. Given three matrices that $\mA \in \mathbb{R}^{M \times K}$, $\mW \in \mathbb{R}^{K \times N}$, and $\mC=\mA \mW \in \mathbb{R}^{M \times N}$, 
and $\boldsymbol{\Theta} \in \mathbb{Z}^{K\times (S/2)}$ is the compression of $\W$, the matrices $\mA$, $\boldsymbol{\Theta}$, and $\mC$ are partitioned into blocks ($\mA_{i,k}$, $\boldsymbol{\Theta}_{k,j}$, and $\mC_{i,j}$) according to the following formulas:
\begin{equation}
\left\{
\begin{aligned}
\mA_{i,k} &= \mA[i \cdot R - (R-1) : i \cdot R, \, k \cdot T-(T-1) : k \cdot T] \\
\boldsymbol{\Theta}_{k,j} &= \boldsymbol{\Theta}[k \cdot T - (T-1) : k \cdot T, \, j \cdot \frac{S}{2}-(\frac{S}{2}-1) : j \cdot \frac{S}{2}] \\
\mC_{i,j} &= \mC[i \cdot R - (R-1) : i \cdot R, \, j \cdot S-(S-1) : j \cdot S]
\end{aligned},
\right.
\end{equation}
where $\mA_{i,k} \in \mathbb{R}^{R \times T}$, $\boldsymbol{\Theta}_{k,j} \in \mathbb{Z}^{K \times (S/2)}$, $\mC_{i,j} \in \mathbb{R}^{R \times S}$. $i \in \{ 1, \cdots, M/R \}$, $j \in \{ 1, \cdots, N/S \}$, $k \in \{1, \cdots, K/T \}$. 
As shown in Figure \ref{HyperLinear_fig}, the decompression can be fused with matrix product as follows:
\begin{equation}
\left\{
\begin{aligned}
&\mS = \frac{l}{l + \frac{\lfloor \mathbf{\Theta_{k,j}} / U \rfloor}{M} \cdot (2l_f - l)} \\
&\a = \frac{[\frac{l}{U},l]}{\sqrt{(\frac{l}{U})^2+l^2}} \\
& \mC_{i,j}= \\
&\sum_{k=1}^{K/T} \mA_{i,k} \cdot (\tau'(\boldsymbol{\Theta}_{k,j} \otimes \a)-1_{K \times\frac{S}{2}} \otimes [\bar x_i, \bar y_i])/\mS + \\
&~~~~~~~~1_{K \times\frac{S}{2}} \otimes [\bar x_i, \bar y_i] \\
\end{aligned},
\right.
\label{eq_birkhoff}
\end{equation}
where we denote $1_{K \times\frac{S}{2}}$ as a matrix with all elements equal to 1 for simple representation in the formula. We term Eq. \eqref{eq_birkhoff} as HyperLinear.

\section{Experiments} \label{experiments}
In this section, we extensively evaluate \textit{Birkhoff} across a broad range of models. i) First, we describe the details of the experimental settings, including descriptions of datasets, hyperparameter configurations of our \textit{Birkhoff} compression framework for each model, and segmentation prompt settings. ii) Second, we conduct a comprehensive performance evaluation of \textit{Birkhoff} on 18 models, including SAM and its variants, measuring compression time, compression ratio, and inference time. We then compare pre-compression and post-compression performance of 18 models on three datasets (COCO, LVIS, and SA-1B) to demonstrate the superiority of our method in preserving model performance. iii) Finally, we benchmark \textit{Birkhoff} against 8 other model quantization techniques, including PTQ4SAM~\cite{lv2024ptq4sam}, QDrop~\cite{wei2022qdrop}, on the COCO dataset. As a mainstream and well-established approach in model compression, model quantization can achieve state-of-the-art results on numerous models. Accordingly, our experiments focus solely on comparisons with quantization. We emphasize that all the results obtained by \textit{Birkhoff} are achieved in a zero-shot manner, without any dataset used for fine-tuning.

\subsection{Experimental Settings} \label{experimental_settings}

\subsubsection{Datasets Description}
We evaluate the effectiveness of our method in the image segmentation task across three datasets, which are COCO\cite{lin2014microsoft} (Common Objects in Context), LVIS\cite{gupta2019lvis} (Large Vocabulary Instance Segmentation), SA-1B\cite{kirillov2023segment} (Segment Anything 1-Billion), respectively.

\textbf{Image segmentation dataset.} The LVIS dataset contains over 1,200 object categories, many of which are long-tailed. Its design emphasizes both the diversity of object categories and the challenge posed by rare categories, making it a valuable benchmark for evaluating the generalizability of segmentation models. The SA-1B dataset is released alongside SAM. It contains over 1 billion masks across 11 million images, created semi-automatically using a promptable segmentation model. Due to the massive scale of the SA-1B dataset and the absence of a designated test set, we randomly select a small subset for evaluation, comprising 1,146,227 segmentation instances. For LVIS, we use its validation set, including 794,250 segmentation instances.

\textbf{Object detection dataset.} COCO is a dataset used for the object detection task. It contains over 118,000 training images and 5,000 validation images, covering 80 object categories. However, the COCO dataset does not include box prompts. Therefore, we use the object detection results of ViTDet-H~\cite{li2022exploring} and YOLOX~\cite{ge2021yolox} as box prompts for the segmentation of prompted instances, including 92,850 segmentation objects.


\begin{table*}[t]
\centering
\caption{Hyperparameters set for compressing each model with \textit{Birkhoff}.}
\begin{tabular}{c|c|c|c|c} 
  \hline
  \textbf{Model} & \textbf{MAE} & \textbf{$l$} & \textbf{$U$} & \textbf{$M$}\\ 
  \hline
  SAM-B~\cite{kirillov2023segment} & 0.0019 & [0.1] & [1600] & [1,2,3]\\
  SAM-L~\cite{kirillov2023segment} & 0.0019 & [0.1] & [1225, 1600] & [1,2,3]  \\
  SAM-H~\cite{kirillov2023segment} & 0.0019 & [0.1] & [1225, 1600] & [1,2,3]  \\
  SAM-HQ-Tiny~\cite{ke2023segment} & 0.0015 & [0.1] & [1600,2500,3600,6400,8100,10000,22500,40000] & [1,2,3]  \\
  SAM-HQ-B~\cite{ke2023segment} & 0.0020 & [0.1] & [1225,1600] & [1,2,3] \\ 
  SAM-HQ-L~\cite{ke2023segment} & 0.0019 & [0.1] & [1225,1600] & [1,2,3]  \\
  SAM-HQ-H~\cite{ke2023segment} & 0.0019 & [0.1] & [1225,1600] & [1,2,3]  \\
  SAM2-Tiny~\cite{ravi2024sam} & 0.0010 & [0.1] & [1600,2500,3600,6400,8100,10000] & [1,2,3]  \\
  SAM2-S~\cite{ravi2024sam} & 0.0010 & [0.1] & [1600,2500,3600,6400,8100,10000,22500,40000] & [1,2,3]  \\
  SAM2-B~\cite{ravi2024sam} & 0.0010 & [0.1] & [1600,2500,3600,6400,8100,10000,22500] & [1,2,3] \\
  SAM2-L~\cite{ravi2024sam} & 0.0012 & [0.1] & [1600,2500,3600,6400,8100,10000,22500,40000] & [1,2,3] \\
  MobileSAM~\cite{zhang2023faster} & 0.0016 & [0.1] & [1600,2500,3600,6400,8100,10000,22500,40000] & [1,2,3] \\
  MobileSAMv2(ViT-H)~\cite{zhang2023mobilesamv2} & 0.0014 & [0.1] & [1225, 1600] & [1,2,3] \\
  EdgeSAM~\cite{zhou2023edgesam} & 0.0011 & [0.1] & [1600,2500,3600,6400,8100,10000,22500,40000] & [1,2,3] \\
  EdgeSAM-RPN~\cite{zhou2023edgesam} & 0.0011 & [0.1] & [1600,2500,3600,6400,8100,10000,22500,40000] & [1,2,3] \\
  EfficientSAM-Ti~\cite{xiong2024efficientsam} & 0.0013 & [0.1] & [1225,1600] & [1,2,3] \\
  EfficientSAM-S~\cite{xiong2024efficientsam} & 0.0013 & [0.1] & [1225,1600] & [1,2,3] \\
  TinySAM~\cite{shu2025tinysam} & 0.0012 & [0.1] & [1600,2500,3600,6400,8100,10000,22500,40000] & [1,2,3] \\
  \hline
\end{tabular}
\label{tab:settings}
\end{table*}

\subsubsection{Hyperparameters Settings}
As illustrated in Algorithm~\ref{pseudocode}, three hyperparameters: $l$, $U$, and $M$, need to be manually set before compression. Table \ref{tab:settings} summarizes the hyperparameter settings used for each of 18 models evaluated in our experiments. Among them, MAE (Mean Absolute Error) is employed to measure the absolute mean changes in parameters before and after compression. Since our method is data-free, we utilize MAE during the compression process as a simple evaluation for estimating the potential performance degradation. Generally, a smaller MAE indicates a better preservation of model performance. $U$ is typically chosen to be a perfect square, as this helps ensure a more uniform distribution of the sampled trajectory points. Empirically, we find that setting $l = [0.1]$ and $M = [1, 2, 3]$ yields effective compression results across most models.

\subsubsection{Prompts Settings}
We use five types of point prompts for evaluation, Box, Box-Center(BC), Mask-Center(MC), and Mask-Rand(MR). The box prompts used for evaluation are obtained from object detection results produced by YOLOX.

\textbf{Box prompts.} For LVIS and SA-1B, Box prompts are included in the datasets. For COCO, we obtain box prompts from object detection results produced by YOLOX.

\textbf{Box-Center points.} A box is generally represented as $[x, y, l_x, l_y]$, where $[x, y]$ denotes the coordinates of the bottom-left corner in a two-dimensional space. The center point $[X_{bc},Y_{bc}]$ of the box can be computed based on the side lengths $l_x$ and $l_y$ along the x-axis and y-axis as follows:

\begin{equation}
    X_{bc} = \frac{2x+l_x}{2}, 
    \quad
    Y_{bc} = \frac{2y+l_y}{2}.
\end{equation}

\textbf{Mask-Center points.} Let $I[i,j]\in\{0,1\}$ denote the binary segmentation mask over an image of width $W$ and height $H$, where $i=1,\ldots,W$ and $j=1,\ldots,H$ index the horizontal and vertical pixel coordinates, respectively. The Mask-Center point $[X_{mc},Y_{mc}]$ can be calculated as follows:

\begin{equation}
    X_{mc} = \frac{\displaystyle\sum_{i=1}^{W}\sum_{j=1}^{H} i \cdot I[i,j]}
    {\displaystyle\sum_{i=1}^{W}\sum_{j=1}^{H} I[i,j]}\,, 
    \quad
    Y_{mc} = \frac{\displaystyle\sum_{i=1}^{W}\sum_{j=1}^{H} j \cdot I[i,j]}
    {\displaystyle\sum_{i=1}^{W}\sum_{j=1}^{H} I[i,j]}\,.
\end{equation}

\textbf{Mask-Rand points.} A Mask-Rand point is obtained by randomly sampling a pixel from a mask. When a single pixel is sampled, the result is Mask-Rand (1pt.); when two pixels are sampled, it is referred to as Mask-Rand (2pt.).

These diverse types of prompts enable a comprehensive evaluation of segmentation performance and stability for SAM and its variants both before and after compression.

\subsection{Compression Performance of \textit{Birkhoff}} \label{section_perform}

\subsubsection{Compression Time and Ratio}
\begin{figure}[!h]
    \centering
    \includegraphics[width=\linewidth]{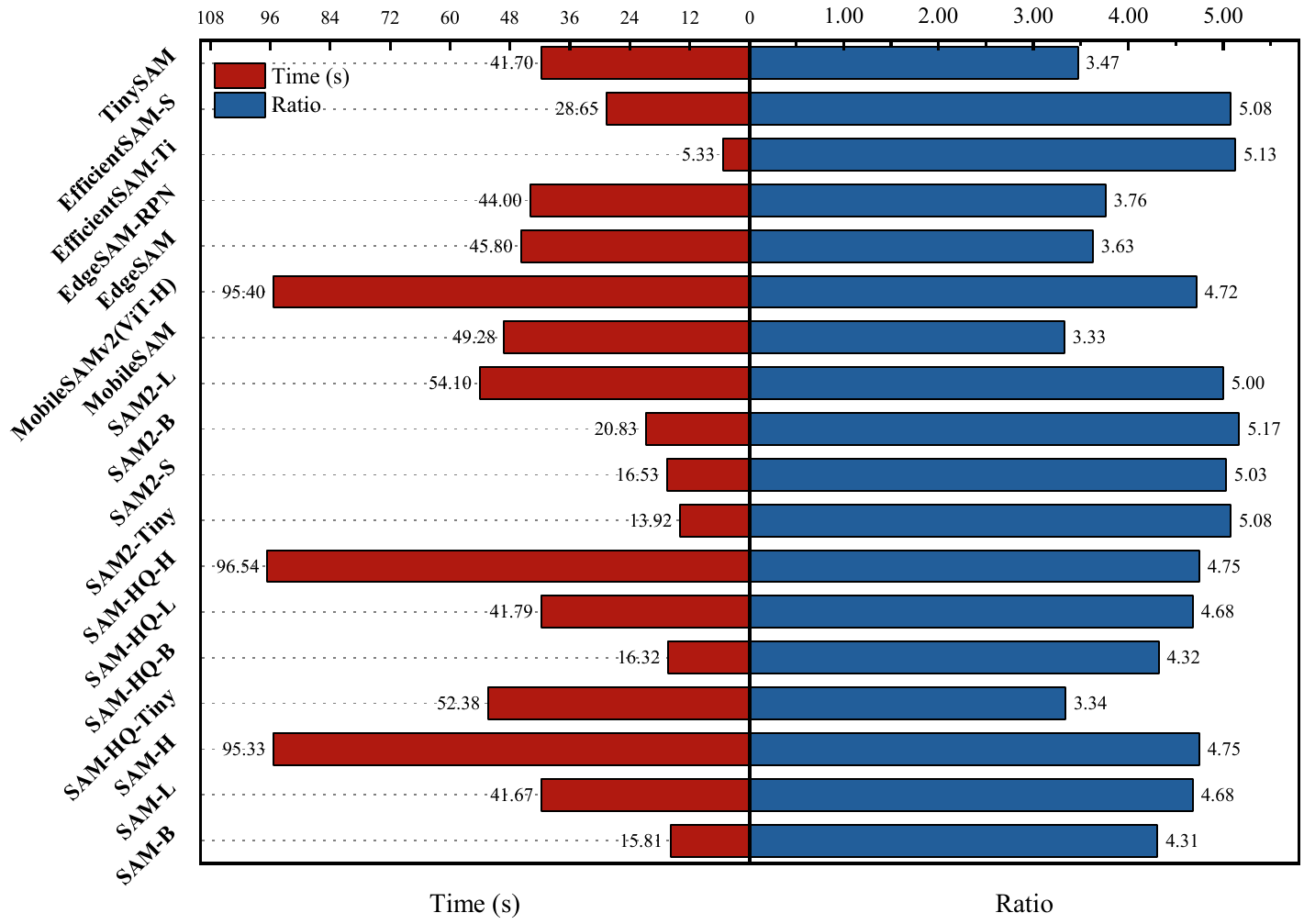}
    \caption{The compression time and achieved compression ratios for 18 SAMs using \textit{Birkhoff}.
}
    \label{tab:compre_time}
\end{figure}

We conduct compression tests on 18 SAMs using the settings in Table~\ref{tab:settings}, reporting both the compression time and achieved compression ratios. As shown in Figure~\ref{tab:compre_time}, our method demonstrates a preferable compression speed and an affordable ratio without both the fine-tuning process and datasets. For instance, SAM-B achieves a 5.17$\times$ ratio in 20.83 seconds, and EfficientSAM-Ti attains 5.13$\times$ compression in 5.33 seconds.

Generally, the compression time increases as the size of the model grows. As shown in Figure~\ref{tab:compre_time}, the majority of models can be fully compressed within 60 seconds. Due to that MobileSAMv2(641M), SAM-HQ-H(643M), and SAM-H(641M) use the ViT-H image encoder, which has more parameters, they require more compression time, taking approximately 90 seconds. However, this is already fast enough for the model compression task. Our advantage is even larger compared to the methods of compression+training and the methods that need to solve an optimization problem.

Small models can be ompressed within 20 seconds. For example, EfficientSAM-Ti(10M) requires only 5.33s, SAM2-Tiny(39M) 13.92s, and SAM-HQ-B(95M) 16.32s. However, some models with small sizes, such as EdgeSAM(10M) and MobileSAM(10M), still require roughly 50 seconds for compression. This is because these models are distilled versions of SAM and thus exhibit a large sensitivity to post-compression parameter loss. Therefore, as shown in Table~\ref{tab:settings}, in order to better preserve model performance, we use a more extensive set of compression hyperparameters, which leads to longer compression times. But we underscore that the overall compression time is still sufficiently fast.

In terms of the compression ratio, most models can be compressed by more than 4.3$\times$, with the highest ratio reaching 5.17$\times$. Even for variants such as EdgeSAM, MobileSAM, and SAM-HQ-Tiny, which have little remaining compressibility after distillation from SAM, our method can still achieve the compression ratios in excess of 3.3$\times$.

\begin{figure*}[!t]
    \centering
    \includegraphics[width=\linewidth]{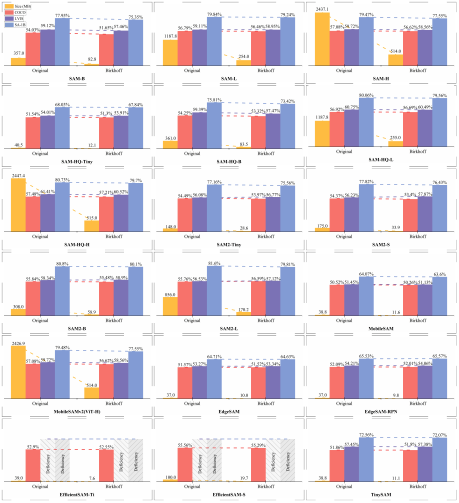}
    \caption{The accuracy and sizes (MB) of 18 SAMs and their corresponding \textit{Birkhoff}-compressed versions on three datasets (COCO, LVIS, and SA-1B).}
    \label{tab:performance}
\end{figure*}

\subsubsection{Preservation of Model Performance}

As shown in Figure~\ref{tab:performance}, we evaluate the accuracy of 18 SAMs and their \textit{Birkhoff}-compressed versions on three datasets (COCO~\cite{lin2014microsoft}, LVIS~\cite{gupta2019lvis} and SA-1B~\cite{kirillov2023segment}) to demonstrate that our method effectively preserves model performance. For the LVIS and SA-1B datasets, we evaluate the mean Intersection over Union (mIoU). For the COCO dataset, to obtain more comprehensive results, we assess not only mIoU but also mask mean Average Precision (mAP) and mean Average Recall (mAR) using boxes as prompts. To produce summary statistics, we average all results for each dataset and compare the performance of the 18 SAMs before and after compression. Detailed experimental results are presented in \textbf{Supplymentary.}

\textbf{Image segmentation results.} As shown in Figure~\ref{tab:performance}, extensive experiments demonstrate that our method effectively preserves the performance of compressed models. 

\textit{In most cases, the change in test scores after compression is within 1\%}. For some models, performance degradation remains within 0.6\%. For example, SAM-L experiences a drop in $0.33\%$ in COCO (from $56.79\%$ to $56.46\%$), $0.16\%$ in LVIS (from $59.11\%$ to $58.95\%$), and $0.60\%$ in SA-1B (from $79.84\%$ to $79.24\%$). Similarly, the results of SAM-HQ-L decrease by $0.26\%$ in LVIS (from $60.75\%$ to $60.49\%$), $0.50\%$ in SA-1B (from $80.06\%$ to $79.56\%$) and $0.23\%$ in COCO (from $56.92\%$ to $56.69\%$). These results are highly favorable, especially considering that they are achieved under data-free conditions in 42 seconds, and at a compression ratio of up to $4.68\times$.

\begin{figure}[ht]
    \centering
    \includegraphics[width=1\linewidth]{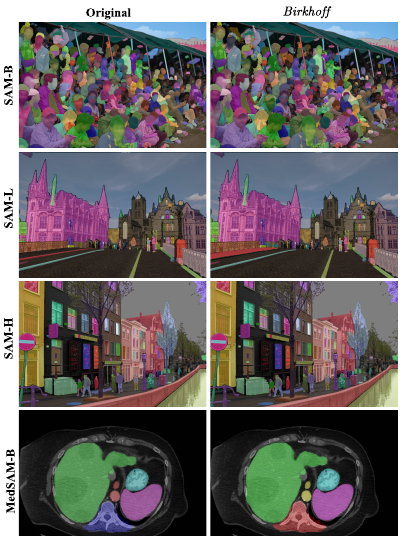}
    \caption{We select SAM-B, SAM-L, SAM-H, and MedSAM-B for a visual comparison of the Segment Everything task before and after \textit{Birkhoff} compression.
}
    \label{segment_every}
\end{figure}

\textbf{Visualization.} As shown in Figure~\ref{segment_every}, using SAM-B, SAM-L, SAM-H, and MedSAM-B as representative models, we conduct comparative experiments on the Segment Everything task with four randomly selected images, including natural and medical images, from the SA-1B dataset and FLARE22 dataset, respectively. Segment Everything refers to simultaneously segmenting all objects within an image. The segmentation results are visualized through color-filled masks. Visualization comparisons indicate that \textit{Birkhoff}-compressed model preserves the boundaries of masks to such an extent that virtually no noticeable visual differences from those of the original models are discernible.

\definecolor{bg01}{HTML}{E2EEFF}
\definecolor{bg02}{HTML}{FFF2CA}
\definecolor{bg03}{HTML}{EDEDED}

\subsubsection{Inference Time}

We evaluate the inference speed of the models before and after replacing the HyperLinear operator. As shown in Table \ref{tab:inference_time}, we conduct comparative experiments on six different models, randomly using a small subset of the SA-1B dataset to measure the average inference time per segmentation. The results indicate that, although the inference speed slightly decreases after incorporating the HyperLinear operator, the performance gap is minimal and barely noticeable to human perception. The data further illustrates that the larger the model, the smaller gap in inference. For example, in the case of SAM-H, the inference speed decreases by only 3.94\%. This is attributed to the nature of our method, where a single-memory access allows the decoding of two parameters, thereby improving the L2 cache hit rate of the operator. For smaller models, most parameters can be accessed directly from registers, ensuring fast memory access. However, for larger models, our method becomes increasingly beneficial by improving the efficiency of memory access. Moreover, we emphasize that though the \textit{Birkhoff}-compressed SAMs still infer more slowly than the original, their differences are at the level of ms, which shall not hurt the user experience.

\begin{table}[ht]
\centering
\caption{Comparison of inference time between the compressed model and the original.}
\vspace{-0.3cm}
\scalebox{1}{
\begin{tabular}{c|c|c} 
  \hline
  \textbf{Model} & \textbf{Original (ms)} & \textbf{\textit{Birkhoff} (ms)}\\  
  \hline
  
  SAM-B~\cite{kirillov2023segment} & 0.314 & 0.388\\
  
  SAM-L~\cite{kirillov2023segment} & 0.721 & 0.809\\
  
  SAM-H\cite{kirillov2023segment} & 1.27 & 1.32\\
  \hline

  SAM-HQ-B\cite{ke2023segment} & 0.317 & 0.395\\ 

  SAM-HQ-L\cite{ke2023segment} & 0.722 & 0.820\\

  SAM-HQ-H\cite{ke2023segment} & 1.26 & 1.35\\
  \hline
\end{tabular}}
\label{tab:inference_time}
\end{table}

\subsection{Comparison with Other Compression Methods} \label{section_comparison}

We compare our \textit{Birkhoff} with both data-free methods, including MinMax\cite{jacob2018quantization}, Percentile\cite{wu2020integer}, and OMSE\cite{choukroun2019low}, and fine-tuning-based methods, such as PTQ4SAM\cite{lv2024ptq4sam}, AdaRound\cite{nagel2020up}, BRECQ\cite{li2021brecq} and QDrop\cite{wei2022qdrop}. Table \ref{tab:yolox} presents a comparison of the mAP scores before and after compressing SAM-B, SAM-L, SAM-H models on the COCO dataset using box prompts.

Maintaining model performance is a critical challenge in model compression. The high-ratio compression becomes meaningless if it induces catastrophic performance degradation. When applied to three SAMs, mAP drops remain within $0.2\%$ (from $37\%$ to $36.8\%$ on SAM-B, from $40.4\%$ to $40.3\%$ on SAM-L, and $41\%$ to $40.8\%$ on SAM-H). Our method realizes near-lossless compression even under data-free constraints. This surpasses all its counterparts. For example, on SAM-B, SAM-L, and SAM-H, we enhance the mAP of OMSE, the best performer, by $23.3\%$, $1.9\%$, and $3.3\%$, respectively. Compared to fine-tuning-based approaches, our method achieves an mAP on SAM-L equivalent to that of PTQ4SAM and surpasses all other methods across three SAM variants. In brief, though the \textit{Birkhoff} is data-free, it is superior to these established quantization methods in compressing SAMs. 

Moreover, our method demonstrates superior stability compared to other data-free methods, exhibiting consistent efficacy regardless of model scale. For instance, OMSE compression causes a $3.5\%$ mAP reduction in SAM-H (from $41.0\%$ to $37.5\%$), but when applied to the smaller SAM-B, performance plummets by $23.5\%$ (from $37.0\%$ to $13.5\%$). In stark contrast, our method consistently maintains deviations below $0.2\%$.

\begin{table}[htb]
    \centering
    \caption{Image segmentation results (mAP) on COCO dataset using box prompts obtained from the objection detection results produced by YOLOX. Methods in blue-shaded columns involve retraining, while those in yellow-shaded columns do not.}
    \scalebox{0.9}{
    \begin{tabular}{c|c|c|c|c}
        \hline
        \textbf{Model} & \textbf{Method} & \textbf{Ratio} & \textbf{Data-Free} & \textbf{mAP (\%)} \\
        \hline
        \multirow{10}{*}{SAM-B\cite{kirillov2023segment}} 
        & \cellcolor{bg01} PTQ4SAM-S-W6A6\cite{lv2024ptq4sam} & \cellcolor{bg01} $\approx 5.33$ & \cellcolor{bg01} No & \cellcolor{bg01} 17.4 \\
        & \cellcolor{bg01} AdaRound-W6A6\cite{nagel2020up} & \cellcolor{bg01} $\approx 5.33$ & \cellcolor{bg01} No & \cellcolor{bg01} 26.4 \\
        & \cellcolor{bg01} BRECQ-W6A6\cite{li2021brecq} & \cellcolor{bg01} $\approx 5.33$ & \cellcolor{bg01} No & \cellcolor{bg01} 26.1 \\
        & \cellcolor{bg01} QDrop-W6A6\cite{wei2022qdrop} & \cellcolor{bg01} $\approx 5.33$ & \cellcolor{bg01} No & \cellcolor{bg01} 33.6 \\
        & \cellcolor{bg01} PTQ4SAM-L-W6A6\cite{lv2024ptq4sam} & \cellcolor{bg01} $\approx 5.33$ & \cellcolor{bg01} No & \cellcolor{bg01} 34.3 \\
        & \cellcolor{bg02} MinMax-W6A6\cite{jacob2018quantization} & \cellcolor{bg02} $\approx 5.33$ & \cellcolor{bg02} Yes & \cellcolor{bg02} 10.7 \\
        & \cellcolor{bg02} Percentile-W6A6\cite{wu2020integer} & \cellcolor{bg02} $\approx 5.33$ & \cellcolor{bg02} Yes & \cellcolor{bg02} 12.0 \\
        & \cellcolor{bg02} OMSE-W6A6\cite{choukroun2019low} & \cellcolor{bg02} $\approx 5.33$ & \cellcolor{bg02} Yes & \cellcolor{bg02} 13.5 \\
        & \cellcolor{bg02} \textbf{Ours(\textit{Birkhoff})} & \cellcolor{bg02} $4.31$ & \cellcolor{bg02} Yes & \cellcolor{bg02} \textbf{36.8} \\
        & \cellcolor{bg03} FP & \cellcolor{bg03} - & \cellcolor{bg03} - & \cellcolor{bg03} 37.0 \\
        \hline
        \multirow{10}{*}{SAM-L\cite{kirillov2023segment}} & \cellcolor{bg01} PTQ4SAM-S-W6A6\cite{lv2024ptq4sam} & \cellcolor{bg01} $\approx 5.33$ & \cellcolor{bg01} No & \cellcolor{bg01} 40.0 \\
        & \cellcolor{bg01} AdaRound-W6A6\cite{nagel2020up} & \cellcolor{bg01} $\approx 5.33$ & \cellcolor{bg01} No & \cellcolor{bg01} 38.9 \\
        & \cellcolor{bg01} BRECQ-W6A6\cite{li2021brecq} & \cellcolor{bg01} $\approx 5.33$ & \cellcolor{bg01} No & \cellcolor{bg01} 38.9 \\
        & \cellcolor{bg01} QDrop-W6A6\cite{wei2022qdrop} & \cellcolor{bg01} $\approx 5.33$ & \cellcolor{bg01} No & \cellcolor{bg01} 39.7 \\
        & \cellcolor{bg01} PTQ4SAM-L-W6A6\cite{lv2024ptq4sam} & \cellcolor{bg01} $\approx 5.33$ & \cellcolor{bg01} No & \cellcolor{bg01} 40.3 \\
        & \cellcolor{bg02} MinMax-W6A6\cite{jacob2018quantization} & \cellcolor{bg02} $\approx 5.33$ & \cellcolor{bg02} Yes & \cellcolor{bg02} 37.5 \\
        & \cellcolor{bg02} Percentile-W6A6\cite{wu2020integer} & \cellcolor{bg02} $\approx 5.33$ & \cellcolor{bg02} Yes & \cellcolor{bg02} 38.0 \\
        & \cellcolor{bg02} OMSE-W6A6\cite{choukroun2019low} & \cellcolor{bg02} $\approx 5.33$ & \cellcolor{bg02} Yes & \cellcolor{bg02} 38.4 \\
        & \cellcolor{bg02} \textbf{Ours(\textit{Birkhoff})} & \cellcolor{bg02} $4.68$ & \cellcolor{bg02} Yes & \cellcolor{bg02} \textbf{40.3} \\
        & \cellcolor{bg03} FP & \cellcolor{bg03} - & \cellcolor{bg03} - & \cellcolor{bg03} 40.4 \\
        \hline
        \multirow{10}{*}{SAM-H\cite{kirillov2023segment}} & \cellcolor{bg01} PTQ4SAM-S-W6A6\cite{lv2024ptq4sam} & \cellcolor{bg01} $\approx 5.33$ & \cellcolor{bg01} No & \cellcolor{bg01} 40.3 \\
        & \cellcolor{bg01} AdaRound-W6A6\cite{nagel2020up} & \cellcolor{bg01} $\approx 5.33$ & \cellcolor{bg01} No & \cellcolor{bg01} 38.3 \\
        & \cellcolor{bg01} BRECQ-W6A6\cite{li2021brecq} & \cellcolor{bg01} $\approx 5.33$ & \cellcolor{bg01} No & \cellcolor{bg01} 38.3 \\
        & \cellcolor{bg01} QDrop-W6A6\cite{wei2022qdrop} & \cellcolor{bg01} $\approx 5.33$ & \cellcolor{bg01} No & \cellcolor{bg01} 40.4 \\
        & \cellcolor{bg01} PTQ4SAM-L-W6A6\cite{lv2024ptq4sam} & \cellcolor{bg01} $\approx 5.33$ & \cellcolor{bg01} No & \cellcolor{bg01} 40.7 \\
        & \cellcolor{bg02} MinMax-W6A6\cite{jacob2018quantization} & \cellcolor{bg02} $\approx 5.33$ & \cellcolor{bg02} Yes & \cellcolor{bg02} 36.1 \\
        & \cellcolor{bg02} Percentile-W6A6\cite{wu2020integer} & \cellcolor{bg02} $\approx 5.33$ & \cellcolor{bg02} Yes & \cellcolor{bg02} 36.3 \\
        & \cellcolor{bg02} OMSE-W6A6\cite{choukroun2019low} & \cellcolor{bg02} $\approx 5.33$ & \cellcolor{bg02} Yes & \cellcolor{bg02} 37.5 \\
        & \cellcolor{bg02} \textbf{Ours(\textit{Birkhoff})} & \cellcolor{bg02} $4.75$ & \cellcolor{bg02} Yes & \cellcolor{bg02} \textbf{40.8} \\
        & \cellcolor{bg03} FP & \cellcolor{bg03} - & \cellcolor{bg03} - & \cellcolor{bg03} 41.0 \\
        \hline
    \end{tabular}}
    \label{tab:yolox}
\end{table}

Therefore, although \textit{Birkhoff} currently does not exceed the compression ratio of the INT6 quantization, it demonstrates compelling merits under stringent constraints: compressing without fine-tuning datasets while rendering almost lossless compression.

\section{Conclusion}
To deploy SAMs efficiently, we propose \textit{Birkhoff}, a universal, data-free, fast, and high-accuracy-compression-ratio model compression algorithm. Not satisfied with translational innovation, our method optimizes Hyper-Compression by introducing a dedicated linear layer operator called HyperLinear to fuse decompression and matrix multiplication to significantly accelerate inference speed of the compressed model. Furthermore, extensive experiments on 18 SAMs in the COCO, LVIS, and SA-1B datasets show that \textit{Birkhoff} performs competitively and consistently in compression time, compression ratio, post-compression performance, and inference speed. In the future, more efforts should be invested in further improving the \textit{Birkhoff} for a better compression ratio surpassing the low-bit quantization.




\bibliographystyle{IEEEtran}
\bibliography{IEEEabrv,reference2}

\vfill

\end{document}